\documentclass[10pt,twocolumn,letterpaper]{article}
\usepackage{xfreqpreprint}

\usepackage{booktabs,multirow,graphicx}
\usepackage{amsmath,amssymb,mathtools}
\usepackage{tikz}
\usetikzlibrary{arrows.meta,positioning,calc}
\definecolor{paperblue}{rgb}{0.21,0.49,0.74}
\usepackage[breaklinks,colorlinks,allcolors=paperblue]{hyperref}
\title{XFreq: Geometry-Aligned Cross-View Residual Learning for
\\
Sparse 3D Ultrasound Reconstruction}
\author{Yuhuan Lu \qquad P.~Ellen Grant \qquad Davood Karimi\\[0.4em]
{\small Department of Radiology, Boston Children's Hospital}\\[-0.1em]
{\small Harvard Medical School, Boston, MA, USA}\\[0.3em]
{\small \href{mailto:yuhuan.lu@childrens.harvard.edu}{\texttt{yuhuan.lu@childrens.harvard.edu}}}}
\begin{document}
\maketitle
\begin{abstract}
Three-dimensional (3D) ultrasound provides valuable volumetric context for anatomical assessment, motivating reconstruction from widely available two-dimensional (2D) slices.
Sparse slices leave much of the volume unobserved, even when their acquisition poses are known.
Existing methods interpolate observed intensities or incorporate learned 2D image features into 3D reconstruction, but their complementary roles in predicting smooth structure and residual details remain less explored.
We introduce XFreq, a geometry-aligned framework for sparse 3D ultrasound reconstruction.
It uses known slice poses and finite footprints to query 2D image features at 3D locations, complementing splatted intensities and sampling geometry.
A low-frequency decoder predicts a smooth 3D field from the splat representation, while a high-frequency decoder additionally incorporates the queried 2D features to predict residual details.
A learned gating module regulates the residual contribution based on geometry and cross-slice feature variation.
We evaluate XFreq on an in-house fetal brain dataset and the public CETUS cardiac dataset, using slices resampled from 3D ultrasound volumes and fivefold cross-validation.
Among the compared methods, XFreq achieves the highest PSNR and SSIM in most settings across both datasets, reaching 26.82\,dB PSNR on CETUS with 32 input slices.
Ablation studies further show the contributions of local feature querying, geometric cues, and residual prediction, with gains generally increasing as more input slices are used.
Our findings highlight the value of geometry-aligned cross-view feature integration for reconstructing 3D ultrasound volumes from sparse 2D slices.
\end{abstract}

\section{Introduction}
\label{sec:intro}
\begin{figure}[t]
\centering
\includegraphics[width=0.45\textwidth]{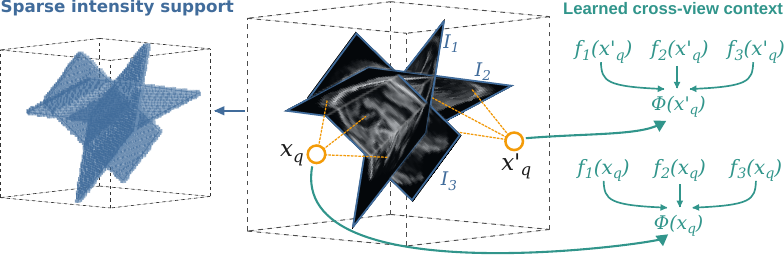}
\caption{Sparse observations and cross-view feature querying.}
\label{fig:problem}
\end{figure}

Ultrasound is widely used in clinical practice, including obstetric, abdominal, and cardiac imaging, because of its affordability, accessibility, and real-time imaging capabilities.
Three-dimensional (3D) ultrasound further provides volumetric anatomical information, enabling examination from different orientations and retrospective selection of imaging planes~\cite{yu20113d}.
As two-dimensional (2D) B-mode ultrasound remains more widely available, reconstructing 3D volumes from sparse 2D slices offers a potential alternative to dedicated 3D acquisition~\cite{iijima2026echopose}.
Standard ultrasound planes are identified by anatomical landmarks, providing spatial cues about the expected relative positions and orientations of 2D views~\cite{cerrolaza20183d,di2023ultrasound,rameshuspose}.
We focus on reconstructing 3D ultrasound volumes from sparse 2D observations with known slice poses.
Traditional methods map observed slice intensities into 3D using known spatial geometry and interpolate unobserved regions~\cite{huang2008mmse,solberg2007freehand}.
Neural implicit and Gaussian representations provide more flexible volumetric models, often optimized separately for each set of input 2D slices~\cite{yeung2024sensorless,eid2025rapidvol,eid2026ultragauss,abdelaziz2026bigus,mildenhall2021nerf}.
Supervised reconstruction methods instead learn common spatial patterns from training data, allowing features extracted from sparse 2D images to guide the prediction of unobserved 3D regions~\cite{lin2023learning,lin2024c,stier2023finerecon}.
Despite these advances, reconstructing unobserved 3D regions from sparse slices remains challenging~\cite{sawdayee2023orex,walker2025crosssdf}, particularly when using learned image features while maintaining consistency with observed slice intensities (Fig.~\ref{fig:problem}).

We introduce \textbf{XFreq}, a geometry-aligned cross-view reconstruction framework that assigns complementary roles to observed intensities and learned image features.
Rather than directly mixing these inputs, XFreq predicts smooth volumetric content from splatted observations and uses pose-aligned image features from multiple slices to infer residual details.
The smooth field and residual are then combined adaptively, followed by a data consistency correction to preserve the observed slice intensities.
To evaluate XFreq across different anatomies and sampling densities, we conduct experiments on fetal brain and cardiac ultrasound datasets using 8, 16, and 32 input slices.
XFreq achieves the highest PSNR and SSIM in most settings among classical, volumetric fitting, and supervised reconstruction methods.
Ablation and spatial analyses further show that the gains from queried image features vary with sampling density and distance to the observed slices.
These findings show that the value of learned image features depends not only on what they encode, but also on how they are spatially aligned with the observed slices.
Our contributions are summarized as follows:
\begin{itemize}
\item We develop a local feature querying mechanism that projects 3D locations onto 2D slices, selects slices based on spatial proximity, and aggregates image features using geometry-conditioned weights.
\item We design an asymmetric architecture that predicts a smooth 3D field from splatted intensities and geometry, while incorporating image features into a gated high-frequency residual branch to recover spatial details.
\item We evaluate XFreq on two ultrasound datasets using fivefold cross-validation.
The results demonstrate the benefits of local feature querying and slice geometry, with queried features providing larger gains near observed slices.
\end{itemize}
\section{Related Work}
\label{sec:related}
Prior work on 3D reconstruction spans classical interpolation, volumetric fitting, and supervised reconstruction.
We review volumetric fitting and learning-based reconstruction, followed by geometric and multiscale techniques for integrating spatial information.

\paragraph{Volumetric Fitting.}
Reconstruction from freehand ultrasound can require estimating unknown slice poses~\cite{dou2024sensorless}.
RecON~\cite{luo2023recon} estimates scan trajectories, while ImplicitVol~\cite{yeung2024sensorless} jointly refines slice poses and fits a continuous intensity field.
Building on implicit volume fitting, RapidVol~\cite{eid2025rapidvol} reduces reconstruction time through factorized volumetric representations.
More recently, 3D Gaussian representations have attracted attention for their explicit spatial parameterization and efficient differentiable rendering~\cite{kerbl20233d}.
UltraGauss~\cite{eid2026ultragauss} models the interaction between acquisition slices and 3D Gaussians, while BIGUS~\cite{abdelaziz2026bigus} and UltraG-Ray~\cite{duelmer2026ultragray} incorporate ultrasound-specific scattering and ray-based acoustic effects.
GaussianPile~\cite{kong2026gaussianpile} further accounts for finite slice thickness in Gaussian-based volumetric reconstruction.
Although these approaches differ in their volumetric representations and image formation models, most optimize a separate 3D representation for each set of input slices, rather than learning a shared reconstruction model from a collection of training volumes.

\paragraph{Learning-Based Reconstruction.}
Learning-based approaches learn a shared reconstruction model from training data to predict 3D volumes from sparse 2D inputs~\cite{jang2025pow3r,charatan2024pixelsplat}.
Normalized convolution~\cite{knutsson1993normalized} and learned confidence propagation~\cite{eldesokey2019confidence} address sparse observations through confidence-aware processing, while 3D U-Net~\cite{cciccek20163d} provides a general architecture for volumetric feature learning.
Learning-based 3D reconstruction in medical imaging includes different prediction targets, ranging from volumetric intensity to anatomical shape~\cite{chen20253d,young2024fully,stevens2025high}.
DIF-Net~\cite{lin2023learning} predicts a continuous 3D intensity field from sparse CBCT projections using learned image features, whereas E-Pix2Vox++~\cite{stojanovski2022efficient} reconstructs 3D cardiac anatomy from multiple 2D ultrasound views.
These studies demonstrate that learned 2D image features can be geometrically aligned and aggregated to guide the reconstruction of unobserved 3D regions~\cite{murez2020atlas,yao2018mvsnet}.

\paragraph{Geometric and Multiscale Techniques.}
Recent methods explore feature consistency and geometric priors to improve 3D reconstruction from sparse slices~\cite{han2025sparserecon,zhang2026anchorsplat}.
FineRecon~\cite{stier2023finerecon} combines depth guidance with high-resolution 2D image features to refine coarse 3D predictions.
C$^2$RV~\cite{lin2024c} combines multiscale volumetric representations with cross-view attention to enhance feature interactions across spatial regions and views.
MS3D~\cite{luo2025ms3d} employs hierarchical multiscale representations to model global structures and local details at different resolutions.
Low-pass filtering and residual decomposition provide a signal processing basis for separating smooth components from fine spatial variations~\cite{mallat1989theory}.
Beyond spatial detail recovery, maintaining consistency with acquired observations is another essential aspect of reconstruction~\cite{liu2023dolce,vo2026efficient}.
MoDL~\cite{aggarwal2018modl} combines learned regularization with explicit data consistency constraints in an iterative reconstruction scheme.
These studies advance both spatial detail recovery and observation consistency, providing a foundation for integrating geometry-aligned image features with directly observed slice intensities in 3D ultrasound reconstruction.

\section{Method}
\label{sec:method}
\begin{figure*}[t]
\centering
\includegraphics[width=1\textwidth]{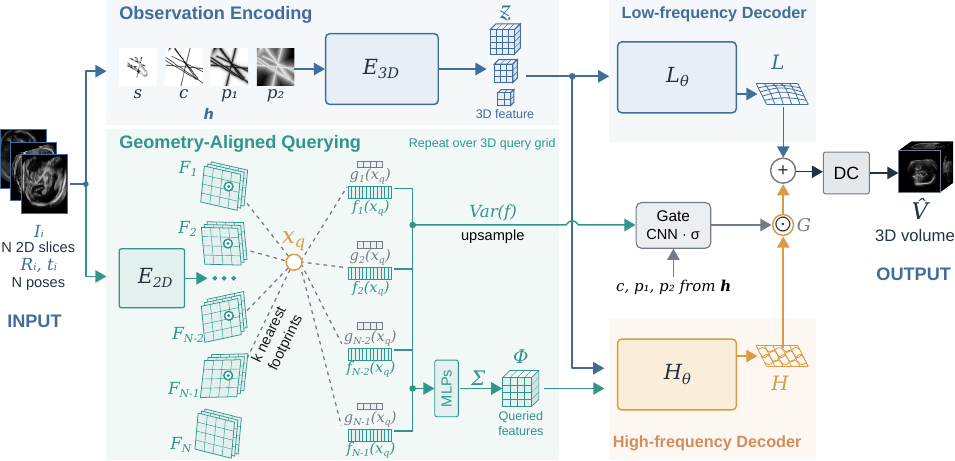}
\caption{\textbf{XFreq architecture.} 
Splatted intensities and pose geometry supply shared 3D features to both decoders. 
Geometry aligned queries aggregate 2D features for the high frequency residual and feature variance for the gate. 
Data consistency is applied after prediction.
}
\label{fig:overview}
\end{figure*}

\subsection{XFreq overview}
\label{sec:method_setting}

Let $\mathcal I=\{(I_i,R_i,t_i)\}_{i=1}^{N}$ denote $N$ observed 2D B-mode ultrasound slices and their known poses, where $I_i\in[0,1]^{H\times W}$ represents the $i$-th slice, and $R_i\in SO(3)$ and $t_i\in\mathbb R^3$ specify its orientation and position in a common 3D coordinate system, respectively.
Our objective is to reconstruct a 3D volume $\widehat V$ from these sparse slices.

As illustrated in Fig.~\ref{fig:overview}, XFreq combines observation encoding and geometry-aligned feature querying for frequency-selective 3D reconstruction.
Splatted slice intensities, sampling coverage, and pose-derived geometry are processed by a shared 3D encoder to obtain multiscale features $\mathcal Z$.
In parallel, a 2D encoder extracts image features from the input slices.
Using known slice poses, XFreq projects each 3D query location onto the slices and aggregates the sampled features into a 3D feature volume $\Phi$.
The low-frequency decoder predicts a smooth 3D field from $\mathcal Z$, while the high-frequency decoder incorporates both $\mathcal Z$ and $\Phi$ to predict residual details.
A learned gate conditioned on sampling geometry and cross-view feature variation modulates the residual before its combination with the smooth prediction, followed by a data consistency correction to produce the final 3D ultrasound volume $\widehat V$.

\subsection{Intensity and Geometry Encoding}
\label{sec:method_observation}
We first construct a 3D representation from splatted slice intensities and sampling geometry, then use a 3D encoder to extract multiscale features.
Preserving the observed intensities is particularly important for patient-specific reconstruction in medical imaging, as individual anatomy may differ from patterns learned across patients.
Let $\mathbf{x}\in\mathbb R^3$ denote a voxel location in the reconstruction grid, with $\widehat V\in[0,1]^{M\times M\times M}$ representing the reconstructed volume.
Using known slice poses, each observed pixel is mapped into 3D, and its intensity is splatted onto neighboring voxels using interpolation weights.
At each voxel location $\mathbf{x}$, $b(\mathbf{x})$ accumulates the weighted pixel intensities, while $w(\mathbf{x})$ accumulates the corresponding interpolation weights.
We construct the splatted intensity and sampling coverage representations $s,c\in\mathbb R^{M\times M\times M}$ as
\begin{equation}
s(\mathbf{x})=\frac{b(\mathbf{x})}{\max(w(\mathbf{x}),\epsilon_s)},\qquad
c(\mathbf{x})=\frac{\log(1+w(\mathbf{x}))}{\kappa},
\label{eq:splats}
\end{equation}
where $\epsilon_s$ ensures numerical stability, the logarithm compresses the range of sampling weights, and $\kappa$ is a fixed normalization constant for coverage scaling.

However, sampling coverage alone does not indicate how far an unobserved voxel is from its nearest input slice.
Let $\Pi_i\subset\mathbb R^3$ denote the finite rectangular footprint of input slice $I_i$ in the common 3D coordinate system.
We therefore define the minimum Euclidean distance from each voxel location to the finite input slice footprints as
\begin{equation}
d_{\min}(\mathbf{x})=
\min_{i=1,\ldots,N}
\min_{\mathbf{r}\in\Pi_i}
\|\mathbf{x}-\mathbf{r}\|_2,
\label{eq:plane_distance}
\end{equation}
where $\mathbf{r}$ represents a point within the slice footprint $\Pi_i$.
Unlike perpendicular distance to an infinite plane, this measure accounts for the finite slice boundaries, including their edges and corners when necessary.
We explicitly encode this minimum distance using two complementary geometry representations $p_1,p_2\in\mathbb R^{M\times M\times M}$:
\begin{equation}
\begin{aligned}
p_1(\mathbf{x}) &=
\exp\left(
-\frac{d_{\min}(\mathbf{x})}{\tau}
\right),\\
p_2(\mathbf{x}) &=
\min\left(
\frac{d_{\min}(\mathbf{x})}{\rho},
\eta
\right),
\end{aligned}
\label{eq:geometry_channels}
\end{equation}
where $\tau$ controls the exponential decay, $\rho$ determines the normalized distance range, and $\eta$ sets the truncation threshold.
The exponential representation $p_1$ emphasizes proximity to the nearest input slice, whereas $p_2$ increases linearly with distance until reaching its upper bound.

Finally, these four spatial representations are stacked into $h=(s,c,p_1,p_2)\in\mathbb R^{4\times M\times M\times M}$.
A multiscale 3D encoder $E_{\mathrm{3D}}$ processes $h$ to obtain hierarchical features $\mathcal Z=E_{\mathrm{3D}}(h)$, which are used by both the low-frequency and high-frequency decoders.

\subsection{Feature Querying and Aggregation}
\label{sec:method_lift}
Observation encoding preserves measured slice intensities and sampling geometry, while learned 2D image features capture structural patterns that can help infer unobserved 3D regions.
For each 3D query location, the known slice poses determine where to sample these features from the corresponding 2D feature maps across views.

A 2D encoder $E_{\mathrm{2D}}$ first extracts a feature map $F_i=E_{\mathrm{2D}}(I_i)\in\mathbb R^{C\times H\times W}$ from each input slice, where $C$ denotes the feature dimension.
To construct a 3D feature representation, we use a query grid of size $M_q\times M_q\times M_q$ that covers the same spatial extent as the reconstructed volume, with $M_q<M$.
For each query location $\mathbf{x}_q\in\mathbb R^3$, we use the known slice poses $(R_i,t_i)$ (Sec.~\ref{sec:method_setting}) to transform it into the local coordinates of the corresponding slice $I_i$:
\begin{equation}
\begin{bmatrix}
u_i(\mathbf{x}_q)\\
v_i(\mathbf{x}_q)\\
\delta_i(\mathbf{x}_q)
\end{bmatrix}
=
R_i^\top(\mathbf{x}_q-t_i),
\label{eq:query_projection}
\end{equation}
where $(u_i,v_i)$ denotes the projected in-plane coordinates and $\delta_i$ is the signed distance along the normal direction to the slice plane.
We obtain the queried feature $f_i(\mathbf{x}_q)\in\mathbb R^C$ by bilinearly sampling $F_i$ at $(u_i,v_i)$.
For projections outside the slice boundaries, feature sampling uses border padding, while an indicator $\chi_i(\mathbf{x}_q)\in\{0,1\}$ records whether the projection lies within the observed slice.

Since the queried features originate from different slice positions and orientations, we incorporate their geometric relationships into the subsequent aggregation.
Unlike the minimum distance encoding in Sec.~\ref{sec:method_observation}, we consider the distances to individual slices for feature aggregation.
Let $d_i(\mathbf{x}_q)$ denote the distance from query location $\mathbf{x}_q$ to the finite footprint $\Pi_i$ of slice $I_i$.
Following Eq.~\ref{eq:geometry_channels}, we obtain slice-specific distance representations $\tilde p_{1,i}(\mathbf{x}_q)$ and $\tilde p_{2,i}(\mathbf{x}_q)$ by replacing $d_{\min}$ with $d_i$ and using $\tau_g$ as the exponential decay parameter.
We further define the slice normal $\mathbf{n}_i=R_i\mathbf{e}_3$, where $\mathbf{e}_3=(0,0,1)^\top$, and let $\bar{\mathbf{n}}$ denote the normalized mean of all the slice normals.

Then we construct a geometric descriptor $g_i(\mathbf{x}_q)\in\mathbb R^8$:
\begin{equation}
\begin{aligned}
g_i(\mathbf{x}_q)=\bigl[&
\frac{\delta_i}{\rho},\,
\frac{|\delta_i|}{\rho},\,
\tilde p_{1,i},\,
\tilde p_{2,i},\\
&
\frac{u_i}{\xi},\,
\frac{v_i}{\xi},\,
\chi_i,\,
1-|\mathbf{n}_i^\top\bar{\mathbf{n}}|
\bigr].
\end{aligned}
\label{eq:geometry}
\end{equation}
The parameter $\rho$ normalizes the distance terms, while $\xi$ normalizes the projected in-plane coordinates.

For each query location $\mathbf{x}_q$, we select the $k$ nearest slice footprints according to $d_i(\mathbf{x}_q)$, with their indices denoted by $\mathcal N_k(\mathbf{x}_q)$.
The queried feature and geometric descriptor are concatenated into $z_i(\mathbf{x}_q)=[f_i(\mathbf{x}_q),g_i(\mathbf{x}_q)]\in\mathbb R^{C+8}$.
Two learned MLPs, $\mathrm{MLP}_{\mathrm{s}}$ and $\mathrm{MLP}_{\mathrm{v}}$, map $z_i(\mathbf{x}_q)$ to a scalar attention score and a $C$-dimensional feature, respectively.
We aggregate the cross-view features as
\begin{equation}
\begin{aligned}
\alpha_i(\mathbf{x}_q)
&=
\frac{\exp\left(\mathrm{MLP}_{\mathrm{s}}(z_i(\mathbf{x}_q))\right)}
{\sum_{j\in\mathcal N_k(\mathbf{x}_q)}
\exp\left(\mathrm{MLP}_{\mathrm{s}}(z_j(\mathbf{x}_q))\right)},\\
\Phi(\mathbf{x}_q)
&=
\sum_{i\in\mathcal N_k(\mathbf{x}_q)}
\alpha_i(\mathbf{x}_q)
\mathrm{MLP}_{\mathrm{v}}(z_i(\mathbf{x}_q)).
\end{aligned}
\label{eq:view_lift}
\end{equation}
$\alpha_i(\mathbf{x}_q)\in(0,1)$ denotes the normalized attention weight assigned to slice $i$, while $\Phi(\mathbf{x}_q)\in\mathbb R^C$ represents the aggregated feature at the query location.
Collecting these features over the query grid yields a 3D feature volume $\Phi\in\mathbb R^{C\times M_q\times M_q\times M_q}$.
This representation combines learned image context from multiple views with the spatial relationships of the observed slices, providing additional information for the high-frequency residual decoder.

\subsection{Frequency-Selective Reconstruction}
\label{sec:method_xfreq}
$\mathcal Z$ encodes observed intensities and sampling geometry, while $\Phi$ represents geometry-aligned 2D image features aggregated across views.
XFreq predicts a smooth intensity field and a complementary residual as the low-frequency and high-frequency components, respectively.
We use $\mathcal Z$ in both branches to ground them in patient-specific observations, while incorporating $\Phi$ only into residual prediction.

To construct training targets for the two branches, we apply a local averaging operator $P$ over a cubic neighborhood of the ground-truth 3D volume $V_{\mathrm{gt}}$ and define
\begin{equation}
V_L^{\mathrm{gt}}=P(V_{\mathrm{gt}}),\quad
V_H^{\mathrm{gt}}=V_{\mathrm{gt}}-V_L^{\mathrm{gt}}.
\label{eq:band_targets}
\end{equation}
The averaging operation uses replicated boundary values to preserve the original volume dimensions.
The low-frequency decoder $L_\theta$ predicts a smooth 3D field $L=L_\theta(\mathcal Z)$, supervised by $V_L^{\mathrm{gt}}$.
In parallel, the high-frequency decoder $H_\theta$ uses $\mathcal Z$ and $\Phi$ to predict a residual $H=H_\theta(\mathcal Z,\Phi)$, whose gated output is supervised by $V_H^{\mathrm{gt}}$.
For residual prediction, $\Phi$ is upsampled as needed and concatenated with the 3D features $\mathcal Z$ at multiple resolutions.

Since the observed slices differ in spatial arrangement and image content (Fig.~\ref{fig:problem}), the contribution of residual details to the final reconstruction may vary throughout the reconstructed volume.
We therefore introduce a learned gate conditioned on sampling geometry and differences among the queried 2D features.
For each query location $\mathbf{x}_q$, we compute the variance of the sampled features across the $k$ nearest slices, averaged over the $C$ feature components:
\begin{equation}
\begin{aligned}
\bar f_j(\mathbf{x}_q)
&=\frac{1}{k}
\sum_{i\in\mathcal N_k(\mathbf{x}_q)}
f_{i,j}(\mathbf{x}_q),\\
a(\mathbf{x}_q)
&=\frac{1}{Ck}
\sum_{j=1}^{C}
\sum_{i\in\mathcal N_k(\mathbf{x}_q)}
\left(f_{i,j}(\mathbf{x}_q)-\bar f_j(\mathbf{x}_q)\right)^2.
\end{aligned}
\label{eq:agreement}
\end{equation}
$f_{i,j}(\mathbf{x}_q)$ denotes the $j$-th feature channel queried from slice $i$, and $\bar f_j(\mathbf{x}_q)$ is its mean across the selected slices.
The resulting $a(\mathbf{x}_q)$ measures feature variance across views, forming a 3D representation $a\in\mathbb R^{M_q\times M_q\times M_q}$ over the query grid.
We upsample $a$ and combine it with sampling coverage $c$ and geometry representations $p_1,p_2$ to predict a voxelwise gate for adaptive residual fusion:
\begin{equation}
G=
\operatorname{sigmoid}\left(
\mathrm{CNN}_{\mathrm{gate}}
\left([c,p_1,p_2,\operatorname{upsample}(a)]\right)
\right),
\label{eq:gate}
\end{equation}
where $\mathrm{CNN}_{\mathrm{gate}}$ denotes a learned 3D convolutional network.
The resulting $G\in(0,1)^{M\times M\times M}$ adaptively regulates the residual contribution at each voxel.

Finally, the network prediction is obtained by adding the smooth intensity field to the residual after voxelwise multiplication with the learned spatially varying gate:
\begin{equation}
\widetilde V=L+G\odot H.
\label{eq:xfreq}
\end{equation}

\subsection{Training and Inference}
\label{sec:method_loss}
To train XFreq, we combine a reconstruction loss with supervision of the smooth and residual predictions.
The total training objective is $\mathcal L=\mathcal L_{\mathrm{rec}}+\mathcal L_{\mathrm{band}}$.

Let $\Omega$ denote the set of voxel locations in the reconstructed volume, and $U=\{\mathbf{x}\in\Omega:w(\mathbf{x})=0\}$ denote the unobserved voxel set.
With the ground truth volume $V_{\mathrm{gt}}$ available, we define the reconstruction error as $e=\widetilde V-V_{\mathrm{gt}}$ and formulate the reconstruction loss as
\begin{equation}
\begin{aligned}
\mathcal L_{\mathrm{rec}}={}&
\frac{1}{2}\operatorname{mean}_{\Omega}(e^2)
+\frac{1}{2}\operatorname{mean}_{U}(e^2)\\
&+\frac{1}{3}\sum_{d=1}^{3}
\operatorname{mean}\left[(\nabla_d e)^2\right],
\end{aligned}
\label{eq:base_loss}
\end{equation}
where $\nabla_d$ denotes the forward finite difference along the $d$-th spatial axis.
The first two terms penalize reconstruction errors across the entire volume and unobserved regions, while the third penalizes spatial gradients of the reconstruction error to encourage the recovery of local details.

To supervise the smooth and residual components defined in Eq.~\ref{eq:band_targets}, we introduce an auxiliary loss:
\begin{equation}
\begin{aligned}
\mathcal L_{\mathrm{band}}={}&
\frac{1}{2}\operatorname{mean}
\left[(L-V_L^{\mathrm{gt}})^2\right]\\
&+\frac{1}{2}\operatorname{mean}
\left[(G\odot H-V_H^{\mathrm{gt}})^2\right].
\end{aligned}
\label{eq:band_loss}
\end{equation}

At inference, XFreq first produces the prediction $\widetilde V$, which is then refined by a data consistency correction to reduce discrepancies with the observed slice intensities.
Let $y$ denote the vectorized intensities of all observed slices, and let $A$ denote the linear operator that samples a 3D volume at the corresponding slice pixel locations.
We estimate a spatially regularized correction field $\delta_{\mathrm{DC}}$ by solving
\begin{equation}
\begin{aligned}
\delta_{\mathrm{DC}}=\arg\min_\delta\quad
&\|A\delta-(y-A\widetilde V)\|_2^2\\
&+\lambda\|D\delta\|_2^2
+\epsilon\|\delta\|_2^2,
\end{aligned}
\label{eq:dc}
\end{equation}
where $D$ denotes a spatial regularization operator, $\lambda$ controls its strength, and $\epsilon$ ensures numerical stability.
We compute the correction field $\delta_{\mathrm{DC}}$ using conjugate gradients and then obtain the final reconstructed volume as
\begin{equation}
\widehat V=
\operatorname{clip}(\widetilde V+\delta_{\mathrm{DC}},0,1).
\label{eq:final_reconstruction}
\end{equation}

\providecommand{\paperTabNum}[2]{\mbox{#1\hspace{.05em}{\tiny$\pm$#2}}}


\begin{table*}[t]
\centering

\caption{\textbf{Comparison of reconstruction methods on \emph{FetalBrain} and \emph{CETUS}.}
Unobserved foreground PSNR (dB) and 3D SSIM are reported as mean $\pm$ standard deviation across five patient folds.
$N$ denotes the number of input slices.
For supervised methods, indented rows labeled ``w/o DC'' report predictions without the shared data consistency correction.}
\label{tab:full_comparison}
\begingroup
\setlength{\tabcolsep}{1.9pt}
\renewcommand{\arraystretch}{0.95}
\scriptsize
\resizebox{\textwidth}{!}{%
\begin{tabular}{@{}l*{12}{c}@{}}
\toprule
\multicolumn{1}{c}{\multirow{3}{*}{\textbf{Method}}}
& \multicolumn{6}{c}{\emph{FetalBrain}}
& \multicolumn{6}{c}{\emph{CETUS}}\\
\cmidrule(lr){2-7}\cmidrule(l){8-13}
& \multicolumn{2}{c}{$N=8$}
& \multicolumn{2}{c}{$N=16$}
& \multicolumn{2}{c}{$N=32$}
& \multicolumn{2}{c}{$N=8$}
& \multicolumn{2}{c}{$N=16$}
& \multicolumn{2}{c}{$N=32$}\\
\cmidrule(lr){2-3}\cmidrule(lr){4-5}\cmidrule(lr){6-7}
\cmidrule(lr){8-9}\cmidrule(lr){10-11}\cmidrule(l){12-13}
& PSNR$\uparrow$ & SSIM$\uparrow$
& PSNR$\uparrow$ & SSIM$\uparrow$
& PSNR$\uparrow$ & SSIM$\uparrow$
& PSNR$\uparrow$ & SSIM$\uparrow$
& PSNR$\uparrow$ & SSIM$\uparrow$
& PSNR$\uparrow$ & SSIM$\uparrow$\\
\midrule

\multicolumn{13}{l}{\emph{Classical Methods}}\\

Interpolation
& \paperTabNum{13.66}{.194} & \paperTabNum{0.386}{.011}
& \paperTabNum{14.56}{.176} & \paperTabNum{0.542}{.005}
& \paperTabNum{15.42}{.140} & \paperTabNum{0.697}{.007}
& \paperTabNum{17.98}{.385} & \paperTabNum{0.530}{.009}
& \paperTabNum{19.48}{.458} & \paperTabNum{0.653}{.012}
& \paperTabNum{20.62}{.518} & \paperTabNum{0.760}{.012}\\

Harmonic
& \paperTabNum{15.17}{.203} & \paperTabNum{0.417}{.010}
& \paperTabNum{16.27}{.174} & \paperTabNum{0.580}{.003}
& \paperTabNum{17.42}{.126} & \paperTabNum{0.733}{.006}
& \paperTabNum{19.84}{.376} & \paperTabNum{0.594}{.015}
& \paperTabNum{21.69}{.469} & \paperTabNum{0.717}{.014}
& \paperTabNum{23.25}{.542} & \paperTabNum{0.810}{.012}\\

NConv~\cite{knutsson1993normalized}
& \paperTabNum{15.41}{.193} & \paperTabNum{0.434}{.009}
& \paperTabNum{16.38}{.150} & \paperTabNum{0.586}{.003}
& \paperTabNum{17.41}{.120} & \paperTabNum{0.722}{.008}
& \paperTabNum{20.07}{.357} & \paperTabNum{0.594}{.011}
& \paperTabNum{21.59}{.436} & \paperTabNum{0.711}{.012}
& \paperTabNum{22.92}{.496} & \paperTabNum{0.801}{.011}\\

\addlinespace[1pt]
\multicolumn{13}{l}{\emph{Volumetric Fitting}}\\

SIREN~\cite{sitzmann2020implicit}
& \paperTabNum{14.92}{.264} & \paperTabNum{0.355}{.016}
& \paperTabNum{17.10}{.224} & \paperTabNum{0.605}{.006}
& \paperTabNum{19.48}{.139} & \paperTabNum{0.786}{.010}
& \paperTabNum{17.59}{.615} & \paperTabNum{0.511}{.014}
& \paperTabNum{20.98}{.627} & \paperTabNum{0.677}{.012}
& \paperTabNum{24.13}{.485} & \paperTabNum{0.809}{.006}\\

ImplicitVol-FP~\cite{yeung2024sensorless}
& \paperTabNum{12.06}{.150} & \paperTabNum{0.084}{.005}
& \paperTabNum{12.71}{.246} & \paperTabNum{0.173}{.015}
& \paperTabNum{13.55}{.294} & \paperTabNum{0.307}{.045}
& \paperTabNum{15.49}{.525} & \paperTabNum{0.271}{.017}
& \paperTabNum{18.31}{.480} & \paperTabNum{0.504}{.017}
& \paperTabNum{21.02}{.582} & \paperTabNum{0.690}{.045}\\

HashGrid~\cite{muller2022instant}
& \paperTabNum{13.39}{.155} & \paperTabNum{0.243}{.005}
& \paperTabNum{13.92}{.245} & \paperTabNum{0.350}{.009}
& \paperTabNum{14.74}{.242} & \paperTabNum{0.522}{.004}
& \paperTabNum{14.15}{.750} & \paperTabNum{0.326}{.025}
& \paperTabNum{15.06}{.742} & \paperTabNum{0.430}{.031}
& \paperTabNum{17.55}{.784} & \paperTabNum{0.624}{.021}\\

TriPlane~\cite{chan2022efficient}
& \paperTabNum{12.63}{.163} & \paperTabNum{0.169}{.007}
& \paperTabNum{13.07}{.189} & \paperTabNum{0.231}{.003}
& \paperTabNum{13.34}{.165} & \paperTabNum{0.281}{.008}
& \paperTabNum{15.38}{.409} & \paperTabNum{0.304}{.007}
& \paperTabNum{17.22}{.405} & \paperTabNum{0.421}{.006}
& \paperTabNum{18.71}{.501} & \paperTabNum{0.525}{.010}\\

BIGUS~\cite{abdelaziz2026bigus}
& \paperTabNum{15.11}{.268} & \paperTabNum{0.453}{.015}
& \paperTabNum{16.99}{.379} & \paperTabNum{0.629}{.021}
& \paperTabNum{15.75}{.934} & \paperTabNum{0.492}{.085}
& \paperTabNum{16.87}{.629} & \paperTabNum{0.533}{.031}
& \paperTabNum{19.04}{.860} & \paperTabNum{0.646}{.017}
& \paperTabNum{19.92}{1.227} & \paperTabNum{0.678}{.042}\\

UltraGauss~\cite{eid2026ultragauss}
& \paperTabNum{15.30}{.339} & \paperTabNum{0.394}{.006}
& \paperTabNum{16.69}{.182} & \paperTabNum{0.583}{.009}
& \paperTabNum{18.22}{.123} & \paperTabNum{0.758}{.007}
& \paperTabNum{20.41}{.371} & \paperTabNum{0.591}{.006}
& \paperTabNum{22.49}{.371} & \paperTabNum{0.721}{.006}
& \paperTabNum{24.66}{.373} & \paperTabNum{0.824}{.006}\\

\addlinespace[1pt]
\multicolumn{13}{l}{\emph{Supervised Methods}}\\

NConv-CNN-3D~\cite{eldesokey2019confidence}
& \paperTabNum{14.52}{.250} & \paperTabNum{0.331}{.017}
& \paperTabNum{15.38}{.342} & \paperTabNum{0.491}{.021}
& \paperTabNum{16.38}{.391} & \paperTabNum{0.671}{.017}
& \paperTabNum{17.25}{.557} & \paperTabNum{0.457}{.012}
& \paperTabNum{19.23}{.602} & \paperTabNum{0.593}{.015}
& \paperTabNum{21.04}{.605} & \paperTabNum{0.714}{.014}\\

\quad\textcolor{black!65}{\emph{w/o DC}}
& \textcolor{black!65}{\paperTabNum{13.66}{.324}}
& \textcolor{black!65}{\paperTabNum{0.120}{.021}}
& \textcolor{black!65}{\paperTabNum{13.92}{.308}}
& \textcolor{black!65}{\paperTabNum{0.146}{.024}}
& \textcolor{black!65}{\paperTabNum{13.96}{.401}}
& \textcolor{black!65}{\paperTabNum{0.175}{.016}}
& \textcolor{black!65}{\paperTabNum{15.36}{.601}}
& \textcolor{black!65}{\paperTabNum{0.268}{.010}}
& \textcolor{black!65}{\paperTabNum{16.20}{.647}}
& \textcolor{black!65}{\paperTabNum{0.307}{.017}}
& \textcolor{black!65}{\paperTabNum{16.59}{.718}}
& \textcolor{black!65}{\paperTabNum{0.333}{.020}}\\

DIF-Net-US~\cite{lin2023learning}
& \paperTabNum{15.47}{.274} & \paperTabNum{0.430}{.016}
& \paperTabNum{16.47}{.287} & \paperTabNum{0.583}{.012}
& \paperTabNum{17.56}{.236} & \paperTabNum{0.739}{.011}
& \paperTabNum{19.09}{.465} & \paperTabNum{0.578}{.012}
& \paperTabNum{21.06}{.584} & \paperTabNum{0.702}{.011}
& \paperTabNum{22.76}{.623} & \paperTabNum{0.803}{.010}\\

\quad\textcolor{black!65}{\emph{w/o DC}}
& \textcolor{black!65}{\paperTabNum{14.98}{.333}}
& \textcolor{black!65}{\paperTabNum{0.296}{.020}}
& \textcolor{black!65}{\paperTabNum{15.64}{.203}}
& \textcolor{black!65}{\paperTabNum{0.375}{.024}}
& \textcolor{black!65}{\paperTabNum{16.31}{.191}}
& \textcolor{black!65}{\paperTabNum{0.474}{.031}}
& \textcolor{black!65}{\paperTabNum{17.98}{.426}}
& \textcolor{black!65}{\paperTabNum{0.469}{.009}}
& \textcolor{black!65}{\paperTabNum{19.18}{.554}}
& \textcolor{black!65}{\paperTabNum{0.540}{.009}}
& \textcolor{black!65}{\paperTabNum{20.10}{.545}}
& \textcolor{black!65}{\paperTabNum{0.602}{.004}}\\

C$^2$RV-US~\cite{lin2024c}
& \paperTabNum{17.46}{.352} & \paperTabNum{0.613}{.021}
& \paperTabNum{18.75}{.259} & \paperTabNum{0.741}{.010}
& \paperTabNum{20.00}{.214} & \paperTabNum{0.843}{.004}
& \textbf{\paperTabNum{22.53}{.204}} & \textbf{\paperTabNum{0.750}{.010}}
& \paperTabNum{24.81}{.251} & \paperTabNum{0.837}{.007}
& \paperTabNum{26.56}{.225} & \paperTabNum{0.895}{.005}\\

\quad\textcolor{black!65}{\emph{w/o DC}}
& \textcolor{black!65}{\paperTabNum{17.44}{.360}}
& \textcolor{black!65}{\paperTabNum{0.605}{.022}}
& \textcolor{black!65}{\paperTabNum{18.71}{.279}}
& \textcolor{black!65}{\paperTabNum{0.731}{.010}}
& \textcolor{black!65}{\paperTabNum{19.89}{.252}}
& \textcolor{black!65}{\paperTabNum{0.829}{.005}}
& \textcolor{black!65}{\paperTabNum{22.47}{.203}}
& \textcolor{black!65}{\paperTabNum{0.739}{.011}}
& \textcolor{black!65}{\paperTabNum{24.65}{.256}}
& \textcolor{black!65}{\paperTabNum{0.819}{.009}}
& \textcolor{black!65}{\paperTabNum{26.20}{.236}}
& \textcolor{black!65}{\paperTabNum{0.869}{.008}}\\


XFreq (ours)
& \textbf{\paperTabNum{17.52}{.263}} & \textbf{\paperTabNum{0.615}{.019}}
& \textbf{\paperTabNum{19.00}{.150}} & \textbf{\paperTabNum{0.753}{.006}}
& \textbf{\paperTabNum{20.38}{.133}} & \textbf{\paperTabNum{0.853}{.005}}
& \paperTabNum{22.52}{.285} & \paperTabNum{0.749}{.009}
& \textbf{\paperTabNum{24.93}{.350}} & \textbf{\paperTabNum{0.840}{.007}}
& \textbf{\paperTabNum{26.82}{.311}} & \textbf{\paperTabNum{0.900}{.005}}\\

\quad\textcolor{black!65}{\emph{w/o DC}}
& \textcolor{black!65}{\paperTabNum{17.51}{.266}}
& \textcolor{black!65}{\paperTabNum{0.609}{.019}}
& \textcolor{black!65}{\paperTabNum{18.97}{.162}}
& \textcolor{black!65}{\paperTabNum{0.744}{.007}}
& \textcolor{black!65}{\paperTabNum{20.29}{.154}}
& \textcolor{black!65}{\paperTabNum{0.841}{.005}}
& \textcolor{black!65}{\paperTabNum{22.46}{.286}}
& \textcolor{black!65}{\paperTabNum{0.742}{.010}}
& \textcolor{black!65}{\paperTabNum{24.78}{.358}}
& \textcolor{black!65}{\paperTabNum{0.829}{.008}}
& \textcolor{black!65}{\paperTabNum{26.51}{.316}}
& \textcolor{black!65}{\paperTabNum{0.882}{.006}}\\

\bottomrule
\end{tabular}%
}
\endgroup
\end{table*}

\section{Experiments}
\label{sec:experiments}
\subsection{Setup}
\label{sec:experimental_setup}
\paragraph{Datasets.}
We evaluate XFreq on the in-house \emph{FetalBrain} dataset (30 volumes from 29 patients) and the public \emph{CETUS}~\cite{bernard2015standardized} dataset (90 cardiac volumes from 45 patients).
The in-house study received institutional review board approval.
For both datasets, we generate paired 2D observations and 3D reconstruction targets from each preprocessed $160^3$ volume with a nominal voxel spacing of 0.6\,mm.
Specifically, we sample $N\in\{8,16,32\}$ 2D slices of size $64^2$ at known poses with 1.5\,mm in-plane spacing, and downsample the source volume to $80^3$ with 1.2\,mm spacing as the reconstruction target $V_{\mathrm{gt}}$ to reduce computational cost.
\paragraph{Compared Methods.}
We compare XFreq with three categories of reconstruction methods.
Traditional methods include nearest-neighbor interpolation, harmonic inpainting, and normalized convolution~\cite{knutsson1993normalized}.
Volumetric fitting methods include SIREN~\cite{sitzmann2020implicit}, ImplicitVol-FP~\cite{yeung2024sensorless}, HashGrid~\cite{muller2022instant}, TriPlane~\cite{chan2022efficient}, BIGUS~\cite{abdelaziz2026bigus}, and UltraGauss~\cite{eid2026ultragauss}.
Learning-based methods trained across patients include NConv-CNN-3D~\cite{eldesokey2019confidence}, DIF-Net-US~\cite{lin2023learning}, and C$^2$RV-US~\cite{lin2024c}, which are adapted to the same 3D ultrasound reconstruction task.
All learning-based methods are trained using the same random seed and evaluated with a shared, frozen data consistency correction.
\paragraph{Implementation Details.}
We implement XFreq in PyTorch and train it on NVIDIA L40S GPUs with 48 GB memory.
Training uses full $80^3$ volumes, with one patient sampled per optimization step and losses averaged across that patient's volumes.
We train for 12,000 iterations per fold using AdamW with an initial learning rate of $10^{-3}$, weight decay of $10^{-4}$, and cosine learning rate decay, alternating among $N=8,16,32$ input slices.
For each fold, we select the checkpoint with the highest mean PSNR on an inner validation set (20\% of the training patients), excluding the test fold.
Additional hyperparameter settings, slice-pose sampling protocols, and data-consistency settings are provided in the supplementary material.
\paragraph{Evaluation.}
We perform fivefold cross validation at the patient level, with each fold serving as the test set in turn.
Reconstruction quality is evaluated using PSNR and 3D SSIM on unobserved foreground voxels, where $w(\mathbf{x})=0$ indicates no direct sampling and $V_{\mathrm{gt}}(\mathbf{x})>5/255$ excludes near-zero background.
Each volume is evaluated using three different sets of slice poses, and the resulting scores are averaged per patient. 
We report the mean and standard deviation across the five folds. 
All methods use identical input slices, poses, and evaluation regions.

\begin{figure*}[t]
    \centering
    \includegraphics[width=\textwidth]{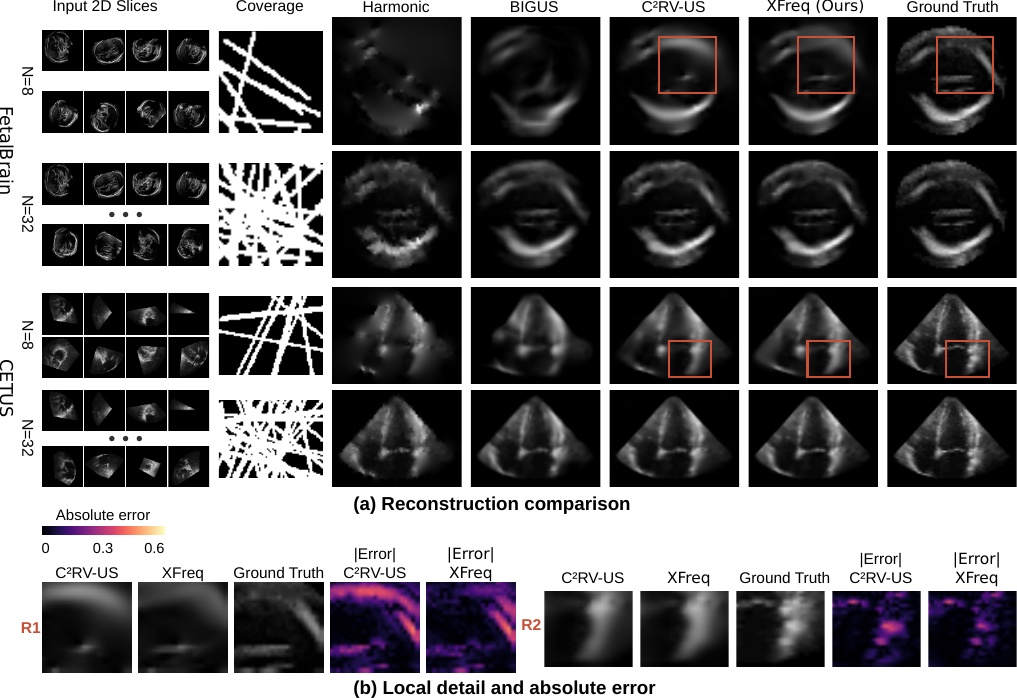}
    \caption{\textbf{Qualitative comparison of 3D ultrasound reconstruction.}
    (a) Reconstruction results on \emph{FetalBrain} and \emph{CETUS} with 8 and 32 input slices.
    (b) Enlarged regions comparing reconstruction details and absolute errors.}
    \label{fig:qualitative}
\end{figure*}

\subsection{Reconstruction Comparison}
\label{sec:result_main}

\paragraph{Quantitative Analysis.}
Table~\ref{tab:full_comparison} presents the reconstruction results of XFreq and representative classical, per-volume fitting, and supervised methods on \emph{FetalBrain} and \emph{CETUS}.
XFreq achieves the highest PSNR and 3D SSIM in most evaluation settings, demonstrating strong reconstruction performance across anatomical structures and sampling densities.
On \emph{CETUS} with $N=8$, C$^2$RV-US slightly exceeds XFreq by 0.01\,dB in PSNR and 0.001 in SSIM.

Increasing the number of input slices generally improves reconstruction accuracy across the evaluated methods.
However, BIGUS~\cite{abdelaziz2026bigus}, which relies on structure-aware Gaussian initialization and error-guided densification, shows non-monotonic performance on \emph{FetalBrain}, with PSNR decreasing from $N=16$ to $N=32$.
Among supervised methods, XFreq achieves the largest improvements, gaining 2.86\,dB on \emph{FetalBrain} and 4.30\,dB on \emph{CETUS}.
Compared with C$^2$RV-US, the PSNR advantage of XFreq increases from 0.06 to 0.25 and 0.38\,dB on \emph{FetalBrain} as $N$ increases.
A similar increasing trend is observed on \emph{CETUS}.
These results suggest XFreq makes more effective use of input slices than C$^2$RV-US, consistent with its geometry-aligned cross-view feature integration design.

To assess the effect of data consistency (DC), we compare supervised methods with and without the same frozen correction in Table~\ref{tab:full_comparison}.
DC improves both PSNR and SSIM across the two datasets and all sampling densities, with larger gains for NConv-CNN-3D and DIF-Net-US than for XFreq and C$^2$RV-US.
As these metrics are evaluated on unobserved voxels, the improvements indicate that the spatially regularized correction benefits regions beyond the observed slices.
Even without DC, XFreq outperforms C$^2$RV-US at $N=32$ by 0.40 and 0.31\,dB on \emph{FetalBrain} and \emph{CETUS}, respectively.
These results suggest that DC provides a useful refinement across methods, while the performance advantage of XFreq is already present before correction.

\paragraph{Qualitative Analysis.}
Figure~\ref{fig:qualitative} compares XFreq with representative methods from three categories: classical reconstruction (Harmonic), volumetric fitting (BIGUS), and supervised learning (C$^2$RV-US).
The figure shows input slices, sampling coverage, and reconstructions alongside ground truth.
All methods are visualized at identical spatial locations for direct comparison.

Compared with Harmonic and BIGUS, XFreq better preserves local intensity variations and structural details in the illustrated examples.
The differences between XFreq and C$^2$RV-US are more subtle, but the enlarged regions show improved detail preservation and lower absolute errors for XFreq in the selected areas.
These observations complement the quantitative results by highlighting the improvements in local reconstruction quality under sparse sampling.
\begin{table*}[t]
\centering
\caption{\textbf{Ablation studies.}
We consider three aspects of XFreq: reconstruction branches and fusion, queried image features, and geometry design.
$\Delta$ denotes XFreq minus the corresponding ablation variant.}
\label{tab:ablation_merged}
\label{tab:branch_ablation}
\label{tab:feature_ablation}
\label{tab:geometry_ablation}
\begingroup
\setlength{\tabcolsep}{1.9pt}
\renewcommand{\arraystretch}{0.95}
\scriptsize
\resizebox{\textwidth}{!}{
\begin{tabular}{@{}l*{12}{c}@{}}
\toprule
\multicolumn{1}{c}{\multirow{3}{*}{\textbf{Configuration}}}
& \multicolumn{6}{c}{\emph{FetalBrain}}
& \multicolumn{6}{c}{\emph{CETUS}}\\
\cmidrule(lr){2-7}\cmidrule(l){8-13}
& \multicolumn{2}{c}{$N=8$}
& \multicolumn{2}{c}{$N=16$}
& \multicolumn{2}{c}{$N=32$}
& \multicolumn{2}{c}{$N=8$}
& \multicolumn{2}{c}{$N=16$}
& \multicolumn{2}{c}{$N=32$}\\
\cmidrule(lr){2-3}\cmidrule(lr){4-5}\cmidrule(lr){6-7}
\cmidrule(lr){8-9}\cmidrule(lr){10-11}\cmidrule(l){12-13}
 & $\Delta$PSNR$\uparrow$ & $\Delta$SSIM$\uparrow$ & $\Delta$PSNR$\uparrow$ & $\Delta$SSIM$\uparrow$ & $\Delta$PSNR$\uparrow$ & $\Delta$SSIM$\uparrow$ & $\Delta$PSNR$\uparrow$ & $\Delta$SSIM$\uparrow$ & $\Delta$PSNR$\uparrow$ & $\Delta$SSIM$\uparrow$ & $\Delta$PSNR$\uparrow$ & $\Delta$SSIM$\uparrow$\\
\midrule
\multicolumn{13}{l}{\emph{Branches and Fusion}}\\
w/o $G\odot H$
& \paperTabNum{+1.05}{.187}
& \paperTabNum{+0.107}{.008}
& \paperTabNum{+1.70}{.165}
& \paperTabNum{+0.112}{.004}
& \paperTabNum{+2.36}{.161}
& \paperTabNum{+0.091}{.009}
& \paperTabNum{+0.82}{.152}
& \paperTabNum{+0.042}{.005}
& \paperTabNum{+1.52}{.266}
& \paperTabNum{+0.046}{.006}
& \paperTabNum{+2.26}{.424}
& \paperTabNum{+0.041}{.007}\\

w/o $G$
& \paperTabNum{+0.02}{.070}
& \paperTabNum{+0.002}{.004}
& \paperTabNum{+0.05}{.063}
& \paperTabNum{+0.003}{.004}
& \paperTabNum{+0.04}{.077}
& \paperTabNum{+0.002}{.002}
& \paperTabNum{+0.08}{.015}
& \paperTabNum{+0.002}{.001}
& \paperTabNum{+0.07}{.053}
& \paperTabNum{+0.001}{.001}
& \paperTabNum{+0.09}{.058}
& \paperTabNum{+0.001}{.001}\\

\addlinespace[1pt]
\multicolumn{13}{l}{\emph{Queried Image Features}}\\
w/o Queried Features
& \paperTabNum{-0.01}{.037}
& \paperTabNum{-0.002}{.004}
& \paperTabNum{+0.04}{.052}
& \paperTabNum{+0.001}{.003}
& \paperTabNum{+0.09}{.045}
& \paperTabNum{+0.002}{.001}
& \paperTabNum{+0.19}{.048}
& \paperTabNum{+0.009}{.002}
& \paperTabNum{+0.30}{.038}
& \paperTabNum{+0.010}{.001}
& \paperTabNum{+0.45}{.060}
& \paperTabNum{+0.008}{.001}\\

w/o Local Querying
& \paperTabNum{-0.01}{.036}
& \paperTabNum{-0.002}{.004}
& \paperTabNum{+0.02}{.048}
& \paperTabNum{+0.001}{.003}
& \paperTabNum{+0.09}{.069}
& \paperTabNum{+0.002}{.002}
& \paperTabNum{+0.13}{.075}
& \paperTabNum{+0.007}{.002}
& \paperTabNum{+0.24}{.062}
& \paperTabNum{+0.008}{.001}
& \paperTabNum{+0.39}{.072}
& \paperTabNum{+0.007}{.001}\\

\addlinespace[1pt]
\multicolumn{13}{l}{\emph{Geometry Design}}\\
w/o Footprint Dist.
& \paperTabNum{+0.05}{.041}
& \paperTabNum{+0.003}{.003}
& \paperTabNum{+0.07}{.019}
& \paperTabNum{+0.003}{.002}
& \paperTabNum{+0.09}{.037}
& \paperTabNum{+0.003}{.001}
& \paperTabNum{+0.04}{.047}
& \paperTabNum{+0.002}{.002}
& \paperTabNum{+0.06}{.048}
& \paperTabNum{+0.002}{.001}
& \paperTabNum{+0.10}{.046}
& \paperTabNum{+0.002}{.001}\\

w/o Footprint Dist. \& $\chi_i$
& \paperTabNum{+0.07}{.052}
& \paperTabNum{+0.005}{.003}
& \paperTabNum{+0.09}{.016}
& \paperTabNum{+0.004}{.002}
& \paperTabNum{+0.11}{.044}
& \paperTabNum{+0.003}{.002}
& \paperTabNum{+0.09}{.078}
& \paperTabNum{+0.004}{.003}
& \paperTabNum{+0.10}{.062}
& \paperTabNum{+0.003}{.001}
& \paperTabNum{+0.12}{.044}
& \paperTabNum{+0.002}{.001}\\

\bottomrule
\end{tabular}%
}
\endgroup
\end{table*}

\begin{figure*}[t]
\centering
\includegraphics[width=1\textwidth]{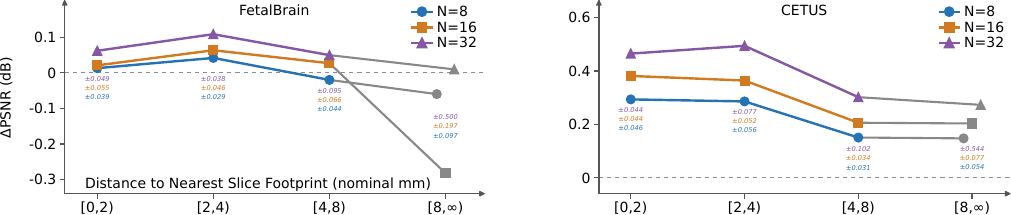}
\caption{\textbf{Spatial distribution of PSNR gains from queried image features.}
Curves show the mean $\Delta$PSNR (dB) of XFreq over the w/o Queried Features variant, while colored labels indicate the corresponding standard deviations across five patient folds.
}
\label{fig:distance_effect}
\end{figure*}

\subsection{Ablation Studies}
\label{sec:result_mechanism}

\paragraph{Reconstruction Branches and Fusion.}
We evaluate reconstruction branches and their fusion through two ablations in Table~\ref{tab:ablation_merged}.
The w/o $G\odot H$ variant retains only $L$ from jointly trained XFreq at inference, while w/o $G$ replaces gated fusion with an unweighted sum $L+H$.
The high-frequency residual improves PSNR and SSIM, with greater PSNR gains at higher sampling densities.
At $N=32$, the PSNR gains reach 2.36 and 2.26\,dB on \emph{FetalBrain} and \emph{CETUS}, respectively.
Compared with direct addition, gated fusion yields additional PSNR gains of 0.04 and 0.09\,dB, respectively.
Within XFreq, adding the residual yields larger gains over $L$ than gating, whose mean improvements are modest and more variable on \emph{FetalBrain}.
\paragraph{Queried Image Features.}
We evaluate queried 2D image features through two ablations in Table~\ref{tab:ablation_merged}.
The w/o Queried Features variant replaces the queried 2D CNN features with zeros before aggregation, while w/o Local Querying replaces them with globally pooled features.
Both variants preserve all other components and training settings.
The performance gains indicate that queried image features improve PSNR and SSIM in most settings.
The PSNR gains increase with the number of input slices on both datasets, suggesting that additional observations provide richer spatial context for feature querying.
At $N=8$, queried features provide no consistent improvement on \emph{FetalBrain}, whereas \emph{CETUS} still benefits from queried features under the same sampling budget.
This difference suggests that the effectiveness of queried features depends on both anatomy and sampling density.
Replacing local queries with globally pooled features generally reduces PSNR, particularly at higher sampling densities, highlighting the importance of spatial correspondence beyond feature content alone.


\paragraph{Spatial Analysis of Queried Features.}
To investigate the spatial contribution of queried image features, we analyze the PSNR gains of XFreq over the w/o Queried Features variant across four distance intervals: $[0,2)$, $[2,4)$, $[4,8)$, and $[8,\infty)$\,mm, measured from the nearest observed slice footprint.
Both models are evaluated on identical unobserved foreground voxels within each interval.
Gray markers indicate the farthest interval, where voxel support can be limited at higher sampling densities.

As shown in Fig.~\ref{fig:distance_effect}, queried image features generally provide larger improvements near the observed slices, particularly on \emph{CETUS}.
At $N=32$, the improvements within 8\,mm range from 0.050 to 0.108\,dB on \emph{FetalBrain} and from 0.302 to 0.494\,dB on \emph{CETUS}.
The gains near observed slices generally increase with the number of input slices, suggesting that additional observations strengthen the contribution of queried features.
Beyond 8\,mm, the mean gains decrease; they are negative for \emph{FetalBrain} at $N=8$ and $N=16$, but remain positive on \emph{CETUS}.
On \emph{FetalBrain} at $N=32$, the farthest interval contains very few evaluation voxels, so its mean should be interpreted cautiously.
These results suggest that queried image features primarily benefit regions close to the observed slices, rather than providing uniform improvements throughout the unobserved volume.

\paragraph{Geometry Design.}
To evaluate the role of finite slice geometry, we compare XFreq with two variants in Table~\ref{tab:ablation_merged}.
The w/o Footprint Dist. variant replaces finite slice footprint distances with distances to infinite planes, while w/o Footprint Dist. \& $\chi_i$ additionally removes the footprint indicator $\chi_i$ in Eq.~\ref{eq:geometry}.
All other components and training settings remain unchanged.
XFreq achieves higher PSNR and SSIM than both variants across both datasets and all sampling densities.
At $N=32$, XFreq achieves PSNR gains of 0.09 and 0.11\,dB over the two variants on \emph{FetalBrain}, respectively, while removing both geometric cues reduces PSNR by 0.12\,dB on \emph{CETUS}.
PSNR gains increase with the number of slices on both datasets.
The second variant shows slightly larger mean PSNR reductions than the first on both \emph{FetalBrain} and \emph{CETUS}.
These findings support the importance of finite slice geometry, though the footprint indicator appears to provide only modest additional gains.

\section{Conclusion}
We presented XFreq, a framework for reconstructing 3D ultrasound volumes from sparse 2D slices with known poses.
XFreq combines splatted intensities and finite slice geometry with locally queried 2D features, using a gated residual decoder to refine volumetric predictions.
Experiments on fetal brain and cardiac ultrasound datasets demonstrate strong reconstruction performance across different numbers of input slices.
Ablation studies show that queried features and finite slice geometry improve reconstruction quality, with queried features providing larger gains near observed slices.
Building on our evaluation with 2D slices resampled from 3D volumes, future work will explore extending XFreq to independently acquired ultrasound scans under varying acquisition conditions and assessing its potential for clinically relevant anatomical applications.
\section*{Acknowledgments}
This research was supported in part by the National Institutes of Health (NIH) under award numbers R01NS128281, R01HD110772, R01EB032708, and EB036945. The content of this publication is solely the responsibility of the authors and does not necessarily represent the official views of the National Institutes of Health.
\section*{LLM Usage Statement}
Anthropic Claude Code assisted with code development and debugging.
OpenAI ChatGPT (GPT-6) assisted with manuscript writing, language polishing, and table formatting. 
The authors reviewed the AI-assisted content and remain responsible for the final work.

{\small
\bibliographystyle{ieeenat_fullname}
\bibliography{main}}
\end{document}